\documentclass[letterpaper, 10 pt, conference]{ieeeconf}  

\IEEEoverridecommandlockouts                              

\usepackage[table, dvipsnames, svgnames, x11names]{xcolor} 
\usepackage{graphics} 
\usepackage{bm}
\usepackage{float} 
\usepackage{stfloats}
\usepackage{caption}
\usepackage{cancel}
\usepackage{subcaption}
\usepackage{widetext}
\usepackage{epsfig} 
\usepackage{mathptmx} 
\usepackage{times} 
\usepackage{booktabs}
\usepackage{amsfonts}
\usepackage{amsmath,amssymb}
\usepackage{colortbl}
\usepackage{adjustbox}
\usepackage{multirow}
\usepackage{marvosym}
\usepackage{pifont}
\usepackage{cite}
\let\labelindent\relax
\usepackage{enumitem}
\usepackage[colorlinks,
            linkcolor=blue,
            anchorcolor=blue,
            citecolor=green,
            urlcolor=Aqua,
            backref=page]{hyperref}
\usepackage{mathrsfs}
\usepackage{algorithmicx,algorithm}
\usepackage{threeparttable}
\usepackage[noend]{algpseudocode}
\usepackage{graphicx}
\DeclareMathAlphabet{\mathcal}{OMS}{cmsy}{m}{n}

\title{\LARGE \bf
(Re)$^2$H2O: Autonomous Driving Scenario Generation via Reversely Regularized Hybrid Offline-and-Online Reinforcement Learning
}

\author{Haoyi Niu$^{1\dag}\textsuperscript{\Letter}$, Kun Ren$^{1\dag}$, Yizhou Xu$^{1}$, Ziyuan Yang$^{1}$, Yichen Lin$^{1}$, Yi Zhang$^{1}$ and Jianming Hu$^{1,2}\textsuperscript{\Letter}$
\thanks{\dag Work done with equal contribution.}
\thanks{$^{1}$Department of Automation, Tsinghua University, Beijing, China. $^{2}$Beijing National Research Center for Information Science and Technology, Beijing, China. \textsuperscript{\Letter}Correspondence to: Jianming Hu and Haoyi Niu {\tt\small hujm@mail.tsinghua.edu.cn, nhy22@mails.tsinghua.edu.cn}}
}

\begin{document}

\maketitle
\pagestyle{empty}

\begin{abstract}
Autonomous driving and its widespread adoption have long held tremendous promise. 
Nevertheless, without a trustworthy and thorough testing procedure, not only does the industry struggle to mass-produce autonomous vehicles (AV), but neither the general public nor policymakers are convinced to accept the innovations. 
Generating safety-critical scenarios that present significant challenges to AV is an essential first step in testing. 
Real-world datasets include naturalistic but overly safe driving behaviors, whereas simulation would allow for unrestricted exploration of diverse and aggressive traffic scenarios. Conversely, higher-dimensional searching space in simulation disables efficient scenario generation without real-world data distribution as implicit constraints.
In order to marry the benefits of both, it seems appealing to learn to generate scenarios from both offline real-world and online simulation data simultaneously. 
Therefore, we tailor a \underline{Re}versely \underline{Re}gularized \underline{H}ybrid \underline{O}ffline-and-\underline{O}nline ((Re)$^2$H2O) Reinforcement Learning recipe to additionally penalize Q-values on real-world data and reward Q-values on simulated data, which ensures the generated scenarios are both varied and adversarial.
Through extensive experiments, our solution proves to produce more risky scenarios than competitive baselines and
it can generalize to work with various autonomous driving models. 
In addition, these generated scenarios are also corroborated to be capable of fine-tuning AV performance.

\end{abstract}

\section{Introduction}
Thanks to enormous advancements made by deep learning techniques, autonomous driving has attracted prohibitive amounts of attention and raised high expectations from both the industry and the general public. 
However, autonomous vehicles have to navigate a wide-open environment that contains countless heterogeneous and long-tailed situations, some of which are constantly outside the agents' scope of experience~\cite{filos2020can}. 
Plus the black-box nature of deep learning, statistically validating the applicability of AV would require billions of miles of road testing~\cite{kalra2016driving}. 
It highlights the necessity of developing stringent and considerate testing pipelines~\cite{li2016intelligence,li2018artificial,li2019parallel,li2020theoretical,feng2021intelligent,zhao2016accelerated,o2018scalable,xusafebench} for AVs to accelerate the on-road or simulation test and raise safety in autonomous driving, which would ultimately have a positive impact on public acceptance of the burgeoning technology. 
To this end, the widely acknowledged challenge is creating safety-critical scenarios~\cite{ding2022survey} in advance that are rarely seen in naturalistic driving data (NDD).

Human driving behaviors in naturally collected datasets are overly conservative and less representative of long-tailed scenarios, sampling from which we would fail to identify failure cases for AVs. 
Picking the outlier scenarios after clustering~\cite{kruber2018unsupervised,wang2018extracting} or applying random perturbation to datasets~\cite{scanlon2021waymo} is insufficient; instead, learning a density model from the data to approximate the scenario distribution~\cite{wheeler2015initial,huang2018synthesis} or directly calculating the distribution~\cite{sun2021corner,feng2021intelligent}, and then sampling events from them according to some types of risky indices~\cite{akagi2019risk}, has fast become an attractive solution. 
It should be noted, however, that the more long-tailed the scenarios are in a dataset, the less accurate the distribution estimation will be, which sometimes enables explicitly modelling NDD distribution a thankless chore.
In addition, sampling rare events from NDD distribution can hardly enhance diversity at root.
To get around these issues, an enticing alternative is searching adversarial background vehicle (BV) policy in unrestricted simulation without processing logged data. 
Having no NDD distribution as constraints on searching space, high dimensionanlity~\cite{feng2021intelligent}, the biggest headache in scenario generation, will be magnified.
With long-term prediction ability to handle temporal complexity, Deep Reinforcement Learning (DRL) attracts new research interests in scenario generation, whereas previous approaches~\cite{koren2018adaptive,koren2019efficient,lee2020adaptive,corso2019adaptive} involve only one or two pedestrians.
This dilemma reminds us of learning adversarial BV policy from offline NDD and online simulation concurrently to address the drawbacks from both, where far too little attention has been paid to date. 

For higher training efficiency~\cite{songhybrid,wagenmaker2022leveraging} and sim-to-real transferability~\cite{niu2022when}, combining offline and online RL has gained fresh prominence with growing number of practical algorithms and insightful motivations. 
We tailor a special recipe out of Hybrid Offline-and-Online RL~\cite{niu2022when} with reverse value regularization, dubbing it (Re)$^2$H2O: penalize Q-values on NDD and boost those on simulation rollouts, so that BVs are encouraged to explore more aggressive behaviors efficiently with NDD distribution as implicit constraints.
We evaluate (Re)$^2$H2O against multiple competing baselines with HighD dataset\cite{krajewski2018highd} in SUMO simulator\cite{behrisch2011sumo} with hand-crafted dynamics model and collision detection mechanism. 
Our solution proves to be capable of generalizing to challenge different AV models by inducing more collisions in a certain time duration and test distance.
Meanwhile, it is notable that fine-tuning on (Re)$^2$H2O scenarios could enhance AV agent performance against applying other generation methods.

\section{Related Work}\label{2}

\subsection{Searching Algorithm for Scenario Generation}\label{2-1}
Many algorithms have been applied to AV critical scenario generation, such as deep generative models~\cite{yang2020surfelgan,chen2021geosim,ehrhardt2020relate,Mhakansson2021}, policy gradient~\cite{chen2021adversarial,ding2020learning,wachi2019failure}, Bayesian optimization~\cite{abeysirigoonawardena2019generating} and evolutionary algorithms~\cite{klischat2019generating}. 
From a methodological standpoint, \cite{ding2022survey} summarized and grouped the current algorithms of safety-critical scenario generation into three types: 
data-driven, adversarial, and knowledge-based generation.

Data-driven scenario generation solely relies on data from previously gathered real-world datasets. 
Aggregating similar scenarios using unsupervised clustering methods~\cite{kruber2018unsupervised,wang2018extracting} can extract rare scenarios from any dataset, but clustering the entire dataset might be ineffective and misleading since scenarios are typically intricate and made up of distinct elements. 
\cite{Ding2018ANM} learns a latent space of encounter trajectories using Variational Auto-Encoder (VAE) and samples unknown scenarios from the latent space, whereas the scheme might be uncontrollable with a limited grasp of the latent code.
This line of work defends the naturality of generated scenarios, but it results in lower efficiency and less diverse scenarios due to the absence of simulation data.

Adversarial-based approaches are to find failure AV cases with unrestricted simulation exploration by constructing adversarial cost function~\cite{abeysirigoonawardena2019generating,wang2021advsim,hanselmann2022king} and often model scenario generation issues as Markov Decision Process (MDP) handled by RL~\cite{koren2018adaptive,ding2020learning,sun2021corner}.
Adaptive Stress Testing (AST)~\cite{koren2019efficient,lee2020adaptive,corso2019adaptive} adopts Monte Carlo tree search (MCTS) and Deep Reinforcement Learning (DRL) to solve scenario generation problems. 
\cite{ding2020learning} uses the combination of different blocks to model the scenarios, and then uses policy gradient RL algorithm to generate safety-critical scenarios.
\cite{karunakaran2020efficient} uses Deep Q-Network to generate the scenarios, which only involves one vehicle and one pedestrian. 
Despite resulting in diverse and fully explored scenarios,
in contrast to data-driven generation, adversarial methods make no attempts to limit the search space, so the generated scenarios can only accommodate a few participants with little complexity.

Knowledge-based approaches leverage external domain knowledge as pre-defined rules~\cite{rana2021building,bagschik2018ontology,menzel2018scenarios} or to perform knowledge-guided learning~\cite{shiroshita2020behaviorally,klischat2019generating,althoff2018automatic}.
In essence, knowledge is formulated as constraints to restrict searching space, while they are not easy to define as is often the case.

\subsection{Leverage of NDD in Scenario Generation}\label{2-2}
While simulation greatly broadens the exploring space and improves scenario diversity, ingenious use of NDD is also essential.
There are many studies focusing on the leverage of NDD in scenario generation~\cite{feng2021intelligent,sun2021corner,wheeler2015initial,huang2018synthesis,suo2021trafficsim,o2018scalable,ding2021multimodal}.

Most existing studies focus on explicitly modeling the distribution of NDD~\cite{wheeler2015initial,wheeler2016factor,suo2021trafficsim,o2018scalable,ding2021multimodal}.
\cite{o2018scalable} uses an adaptive importance sampling method to learn alternative distributions from the base distribution learned from NDD. 
\cite{ding2021multimodal} generates scenarios involving one cyclist to attack the AV. It initializes parameters as a Gaussian distribution and uses NDD distribution as a constraint. 
\cite{knies2020data} extracts useful scenarios according to the cooperative actions for cooperative maneuver planning evaluation. But it focuses on extracting scenarios from existing datasets with no exploration into finding new scenarios.
\cite{Mhakansson2021} introduces recurrent models to generate realistic scenarios of highway lane changes and feeds NDD into the discriminator to improve the generator.

After investigating the previous work mentioned above, we propose a new method 
(Re)$^2$H2O that combines learning from both online simulation and offline NDD. We boost Q-values on simulation rollouts, thus inducing BVs to explore more aggressive behaviors than naturalistic ones. Meanwhile, we penalize Q-values on NDD samples with no need for explicitly recovering NDD distribution and NDD as implicit constraints help overcome high dimensionality issues to some extent. 

\section{Methodology}\label{3}
\subsection{Scenario Formulation}\label{3-1}
A scene at moment $t$ incorporates state $\mathbf{s}_t$ and action vector $\mathbf{a}_t$ to capture the dynamics information of all the vehicles. Considering the traffic involving a test AV ($V_0$) and $N$ BVs ($V_i, i=1,\cdots, N$) driving on the highway, a scene is presented as:
\begin{equation}
\mathbf{s}_t=[\mathbf{s}_t^0,\mathbf{s}_t^1,\cdots,\mathbf{s}_t^N]^{\rm T},\quad \mathbf{u}_t=[\mathbf{a}_t^0,\mathbf{a}_t^1,\cdots,\mathbf{a}_t^N]^{\rm T}\label{s_mat}
\end{equation}
 For each vehicle at moment $t$, the state vector consists of its lateral and longitudinal position $(x, y)$, longitudinal velocity $v$ and heading angle $\theta$, and the action vector comprises the increment of its longitudinal velocity and heading angle between two moments in succession:
\begin{equation}
\mathbf{s}_t^i=[x_t^i,y_t^i,v_t^i,\theta_t^i],\quad \mathbf{a}_t^i=[\Delta v_t^i,\Delta\theta_t^i]
\end{equation}
Borrowing the definition from~\cite{ulbrich2015defining,feng2020testing1}, we depict a scenario as a finite sequence of $H$ frames of traffic scenes:
\begin{equation} 
\mathbf{s}_0\rightarrow \mathbf{u}_0\rightarrow \mathbf{s}_1\rightarrow \mathbf{u}_1\rightarrow \cdots \mathbf{s}_H\rightarrow \mathbf{u}_H
\end{equation}

\subsection{Scenario Generation with Online RL in Simulation}\label{3-2}
Development in a traffic environment can be conceivably constructed as an MDP defined by tuple $\left(\mathcal{S}, \mathcal{A}, \mathcal{R}, P, \rho, \gamma\right)$, where $P$ and $\rho$ are transition probability and initial state distribution respectively.
To obtain safety-critical scenarios, we regard all the BVs as integrated agents trained via online RL in simulation, aiming that BVs can challenge AV with more adversarial \textbf{actions} $\mathbf{a}_t=[\mathbf{a}_t^1,\cdots,\mathbf{a}_t^N]^{\rm T}\in\mathcal{A}$ given \textbf{state} $\mathbf{s}_t\in\mathcal{S}$ to maximize accumulative \textbf{reward} $r_t=r\left(\mathbf{s}_{t}, \mathbf{a}_{t}\right)\in\mathcal{R}$ discounted by $\gamma$: 
\begin{equation}
J(P,\pi)=\mathbb{E}_{\mathbf{s}_0\in\rho,\mathbf{a}_t\sim\pi(\cdot|\mathbf{s}_t), \mathbf{s}_{t+1}\sim P(\cdot|\mathbf{s},\mathbf{a})}\left[\sum_{t=0}^{H} \gamma^{t} r\left(\mathbf{s}_{t}, \mathbf{a}_{t}\right)\right]\label{obj}
\end{equation}
The design of the reward function involves concerns on the distance between AV and BV ($r_t^{dis}$) and collision occurrence ($r_t^{col}$):
\begin{equation}
r_t=r_t^{dis}+r_t^{col}
\end{equation} 
To induce BVs to collide with AV, we reward BVs for any of them being closer to AV with $r_t^{dis}$ in Eq.~\ref{r_ego_adv}, where $d(V_0, V_i),i=1,\cdots,N$ measures the minimum distance between any two points on AV and the $i$-th BV respectively:
\begin{equation}
r_t^{dis}=-\min_i\{\text{d}(V_0, V_i)\},\quad i=1,\cdots,N\label{r_ego_adv}
\end{equation}
While $r_t^{dis}$ performs as a continuous learning signal, agents also deserve a reward $R_b$ for AV collision and a punishment $-R_b$ for BV-to-BV accident at the last moment in the scenario, thus allowing the discovery of AV failure cases still with reasonable cooperative BV behaviors:
\begin{equation}
r_t^{col}=\left\{\begin{aligned}
R_b,& \text{if AV collided with BV}
\\-R_b,& \text{if BV collided with BV}
\\0,& \text{else}
\end{aligned}\right.
\end{equation}
In conventional actor-critic formalism~\cite{sutton1998introduction}, Q-function $\hat{Q}$ is approximated by minimizing the standard Bellman error (\textit{policy evaluation} in Eq.~\ref{pe}), and on the other side, policy $\hat{\pi}$ is optimized by maximizing the Q-function (\textit{policy improvement} in Eq.~\ref{po}):
	\begin{align}
		\hat{Q} &\leftarrow \arg \min _{Q} \mathbb{E}_{\mathbf{s}, \mathbf{a}, \mathbf{s}^{\prime}\sim \mathcal{U}}\left[\frac{1}{2}\left((Q-\hat{\mathcal{B}}^\pi\hat{Q})(\mathbf{s},\mathbf{a})\right)^{2}\right] \label{pe} \\
		\hat{\pi} &\leftarrow \arg \max _{\pi} \mathbb{E}_{\mathbf{s}, \mathbf{a}\sim \mathcal{U}}\left[\hat{Q}(\mathbf{s}, \mathbf{a})\right]\label{po}
	\end{align}
where $\mathcal{U}$ denotes the data buffer generated by a previous version of policy $\hat{\pi}$ through online simulation interactions.
The Bellman evaluation operator $\hat{\mathcal{B}}^{\pi}$ is given by $\hat{\mathcal{B}}^{\pi} \hat{Q}(\mathbf{s}, \mathbf{a})=r(\mathbf{s}, \mathbf{a})+\gamma \mathbb{E}_{\mathbf{a}^{\prime} \sim \hat{\pi}\left(\mathbf{a}^{\prime} | \mathbf{s}^{\prime}\right)}\left[\hat{Q}\left(\mathbf{s}^{\prime}, \mathbf{a}^{\prime}\right)\right]$. For brevity, we denote Bellman error sampling over $\mathcal{U}$ in Eq.~\ref{pe} as $\mathcal{E}_\mathcal{U}(Q,\hat{\mathcal{B}}^\pi\hat{Q})$.

To capture the stochasticity, we model BV policy as Gaussian distribution $\mathcal{N}(\mu,\sigma)$, where $\mu$ and $\sigma$ are approximated with a neural network $\mathcal{W}_\theta$. 
Directly sampling from the Gaussian, $\xi_t$ cannot satisfy real-world requirements and limitations (e.g. velocity limit and maximum steering angle) without $\tanh$ and linear mapping functions.
The estimated policy $\hat{\pi}_\theta(\mathbf{s}_t)$ is therefore implemented by:
\begin{equation}
    \begin{aligned}
    \mu_t, \sigma_t &\leftarrow \mathcal{W}_\theta(\mathbf{s}_t): R^{4\times{(N+1)}}\rightarrow R^2\times R^{2\times N}\\
    \xi_t&\sim\mathcal{N}(\mu_t,\sigma_t): R^2\times R^{2\times N}\rightarrow R^{2\times N}\\
    \zeta_t&\leftarrow \tanh{(\xi_t)}: R^{2\times N}\rightarrow R^{2\times N}\\ 
    \mathbf{a}_t&\leftarrow \textsc{LinearMapping}(\zeta_t): R^{2\times N}\rightarrow R^{2\times N}
    \end{aligned}
\end{equation}
With another neural network, the Q-function can be approached as $\hat{Q}_\phi$. 
Then, iteratively optimizing the Q-function and policy network in Eq.~\ref{pe} and \ref{po} would finally achieve the maximization over objective in Eq.~\ref{obj}, resulting in more aggressive yet still reasonable BV driving models.

\subsection{Incorporate Offline NDD in Online RL}\label{3-3}
On one hand, online RL agents have to search safety-critical scenarios over the whole simulation space, thus bearing low sampling efficiency.
On the other hand, there are large amounts of offline NDD collected by the industry, whereas driving behaviors inside are overly safe and conservative for challenging scenario searching.
The dilemma motivates us to incorporate offline naturalistic data $\mathcal{D}$ in online simulation buffer $B$ to eliminate the naturalistic behaviors and facilitate safety-critical scenario generation, by identifying BV behaviors that substantially depart from the reference of NDD.
To this end, we reverse the previous value regularization scheme in hybrid offline-and-online RL~\cite{niu2022when}, replacing policy evaluation in Eq.~\ref{pe} with the following term:
 \begin{equation}
     \begin{aligned}
	\min_{Q}{\color{Red}\min_{d}}\; \beta\left[\mathbb{E}_{\mathbf{s},\mathbf{a} \sim \mathcal{D}}\left[Q\left(\mathbf{s}, \mathbf{a}\right)\right]-{\color{Red}\mathbb{E}_{\mathbf{s}, \mathbf{a}\sim d(\mathbf{s},\mathbf{a})}\left[Q(\mathbf{s}, \mathbf{a})\right]}\right. \\+ \left.{\color{Red}\mathcal{R}(d)}\right]+\mathcal{E}_{B\cup\mathcal{D}} \left(Q, \hat{\mathcal{B}}^{\pi} \hat{Q}\right)\label{d^phi}
	\end{aligned}
 \end{equation}
where $\beta$ is a positive scaling parameter. $d(\mathbf{s}, \mathbf{a})$ is a particular state-action sampling distribution, and $\mathcal{R}(d)$ is a regularization term for $d$ to impose some type of sampling preference. 
Ideally, we tend to penalize Q-values at the NDD samples and pull up those on simulation samples, especially with higher diversity. 
Therefore, we select $\mathcal{R}(d)=-\mathcal{H}(d)=\mathbb{E}_{\mathbf{s}, \mathbf{a}\sim d(\mathbf{s},\mathbf{a})}\left[\log d(\mathbf{s}, \mathbf{a})\right]$ where $\mathcal{H}(d)$ is computed as entropy, and solve the inner-loop minimization problem that draws $d$ closer to a diverse sampling distribution and meanwhile assigns it higher probabilities to capture samples with larger Q-values, commonly standing for more aggressiveness.

The inner-loop problem (red in Eq.~\ref{d^phi}) can be reformulated as a constrained optimization:
\begin{equation}
\begin{aligned}
    \min_{d}&\ -\mathbb{E}_{\mathbf{s}, \mathbf{a} \sim d(\mathbf{s}, \mathbf{a})}[Q(\mathbf{s}, \mathbf{a})]-\mathcal{H}[d(\mathbf{s}, \mathbf{a})] \\
    \text{s.t.}&\ \iint_{\mathbf{s}, \mathbf{a}} d(\mathbf{s}, \mathbf{a})\mathrm{d}\mathbf{s}\mathrm{d}\mathbf{a}=1, d(\mathbf{s}, \mathbf{a})  \succeq \mathbf{0}
\end{aligned}
\label{closed-form}
\end{equation}
Solving Eq.~\ref{closed-form} derives the closed-form solution $d^\star(\mathbf{s}, \mathbf{a})=\exp[Q(\mathbf{s},\mathbf{a})]/\iint_{\mathbf{s},\mathbf{a}}\exp [Q(\mathbf{s},\mathbf{a})]\mathrm{d}\mathbf{s}\mathrm{d}\mathbf{a}$.
As the denominator is a constant that does not depend on state-action pairs anymore, $d^\star$ is proportional to the soft-maximum Q-values $\exp[Q(\mathbf{s},\mathbf{a})]$, sampling from which echos the demand of diversifying BV policy to achieve higher Q-values and thus more risky scenarios for AV.
Plugging $d^\star$ back into Eq.~\ref{d^phi}, we obtain the final policy evaluation scheme:
\begin{equation}
\begin{aligned}
\min_Q\beta\left[{\mathbb{E}_{\mathbf{s},\mathbf{a}\sim\mathcal{D}}[Q(\mathbf{s},\mathbf{a})]-\log\iint_{\mathbf{s},\mathbf{a}}\exp{[Q(\mathbf{s},\mathbf{a})]}\mathrm{d}\mathbf{s}\mathrm{d}\mathbf{a}}\right]\\+\mathcal{E}_{B\cup\mathcal{D}}(Q,\hat{\mathcal{B}}^\pi\hat{Q})
\end{aligned}\label{new_pe}
\end{equation}
which minimizes the standard Bellman error over $B$ and $\mathcal{D}$, additionally pushing down Q-values on offline naturalistic driving samples with $\mathbb{E}_{\mathbf{s},\mathbf{a}\sim\mathcal{D}}[Q(\mathbf{s},\mathbf{a})]$ and further pulling up Q-values on simulation samples with higher Q-values with \texttt{log-sum-exp}. The integral on $\exp[Q(\mathbf{s},\mathbf{a})]$ relies on sampling over the whole state-action space, which can be impractical and unnecessary since not all state-action pairs are valid in the real world and can be enumerated in the simulation. 
Instead, we sample state-action pairs from the simulation replay buffer $B$ (with sufficient coverage) to approximate the value. Optimizing $\hat{Q}$ with the objective in Eq.~\ref{new_pe} and $\hat{\pi}$ with the following policy improvement term would finally accomplish the risky scenario generation task with the combination of online simulation data $B$ and offline NDD $\mathcal{D}$:
\begin{equation}
    \max _{\pi} \mathbb{E}_{\mathbf{s}, \mathbf{a}\sim B\cup\mathcal{D}}\left[\hat{Q}(\mathbf{s}, \mathbf{a})\right]\label{new_po}
\end{equation}

\section{Experiments}\label{4}
We experimentally demonstrate that (Re)$^2$H2O is an effective approach for risky scenario generation by providing evidence for the following questions as listed:
\begin{itemize}[leftmargin=*,topsep=0pt]
    \item Can (Re)$^2$H2O generate scenarios with more  risky characteristics from naturalistic driving dataset? (Sec.~\ref{NDD})
    \item Can (Re)$^2$H2O generate more risky scenarios and better enhance test efficiency than other baselines? (Sec.~\ref{compare})
    \item Can the performance of learning-based AV model be improved by augmenting training distribution with scenarios generated by (Re)$^2$H2O? (Sec.~\ref{fine})
    \item How does (Re)$^2$H2O regularize Q-values on offline naturalistic driving data and online simulation data? (Sec.~\ref{didactic})
\end{itemize}

\subsection{Experimental Settings}\label{4-1}
\subsubsection{Dataset Processing}\label{4-1-1}
We adopt HighD dataset~\cite{krajewski2018highd} as NDD in this paper. It contains collective naturalistic vehicle trajectories on German expressways, with a total mileage of 44,500 kilometers. 
It also provides rich information about vehicle position, velocity, size, static road environment and macroscopic traffic flow, which is adequate for scenario analysis and reconstruction. 
The reaction interval in simulation for all the experiments is aligned with the sampling interval (0.04 seconds) in HighD data.

In this work, we target traffic flow on three-lane highways and select segments of 2-5 vehicles with arbitrarily one of them as AV and the others as BVs, when all the distances between each other are less than a certain threshold. 
Since not all segments in HighD is of interest for scenario generation, processed vehicle data should satisfy that the velocity is limited in range 0 to 40m/s and the acceleration is limited in range $-0.8g$ to $0.6g$ where $g$ is gravity acceleration. 
After processing, we obtain $3\times 10^5$ data points that can later be built into scenarios. 
Specifically, these data points are also 
standardized as transitions $(\mathbf{s},\mathbf{a},r,\mathbf{s}',\texttt{done})$ that can be leveraged by RL algorithms directly.

\subsubsection{Metrics}\label{4-1-2}
To assess the scenario quality, we initially select three widely acknowledged metrics for evaluations and comparisons. 
\textbf{Collision Rate (CR)} is the percentage of AV-BV collision scenarios out of all the evaluation scenarios. \textbf{Average Collision Time (ACT)} and \textbf{Average Collision Distance (ACD)} represent time duration and distance of AV with BVs before accidents in collision scenarios.

However, low ACT/ACD with low CR indicates that BV often luckily runs into AV with relatively random and imprudent behaviors.
Besides, it is most likely that a method trades longer searching time and larger traveling distance for more successful AV-BV accidents.
Therefore, CR, ACT and ACD might not always give clear evidence for generation methods to produce more risky scenarios, since they are unaware of time consumption and traveling distance that always limit the deployment of AV testing procedure.
To this end, we additionally propose two objective and reasonable metrics: 
\textbf{Average Collision Frequency Per Second (CPS)} and \textbf{Average Collision Frequency Per Meter (CPM)}, 
given by AV-BV collision numbers $N_{col}$ averaged by total testing time $T_{tt}$ and total AV testing distance $D_{tt}$ respectively:
\begin{equation}
\begin{aligned}
    \text{CPS}=\frac{N_{col}}{T_{tt}}, \quad \text{CPM}=\frac{N_{col}}{D_{tt}}
\end{aligned}
\label{CPS and CPM}
\end{equation}


\subsubsection{Baselines}\label{4-1-3}
We design three baselines for comparisons:
\begin{itemize}[leftmargin=*,topsep=0pt]
    \item \textbf{Domain Randomization (DR)}: we randomize every BV a constant velocity in each scenario~\cite{niu2021dr2l}. Plus, BVs can choose to change or keep lane with equal probability at any time during a scenario.
    \item \textbf{(Re)$^2$H2O without NDD (\textit{w/o real})}: 
    set the simulation and real data ratio as $\infty$ in each batch.
    \item \textbf{(Re)$^2$H2O without simulation data (\textit{w/o sim})}: set the simulation and real data ratio as 0 in each batch.
\end{itemize}
To ensure fair comparisons, we set all the hyperparameters in \textit{w/o real} and \textit{w/o sim} consistent with (Re)$^2$H2O unless otherwise specified.


\subsubsection{Test AV Driving Policies}\label{4-1-4}
4 different driving policies are applied on AV for test: 
1) na\"ive \textbf{Uniform Linear Motion}; 
2) \textbf{SUMO
 Car-Following Model}~\cite{song2014research} with SUMO lane-changing model~\cite{erdmann2015sumo};
 3) \textbf{FVDM Car-Following model}~\cite{jiang2001full} with SUMO lane-changing model; 
 4) \textbf{Agent model trained by RL}. 
 For simplicity, we dub these driving policies as ``uniform", ``sumo", ``fvdm" and ``RL-agent" hereinafter. 

\begin{figure}
    \adjustbox{center}{
        \centering
        \includegraphics[width=0.5\textwidth]{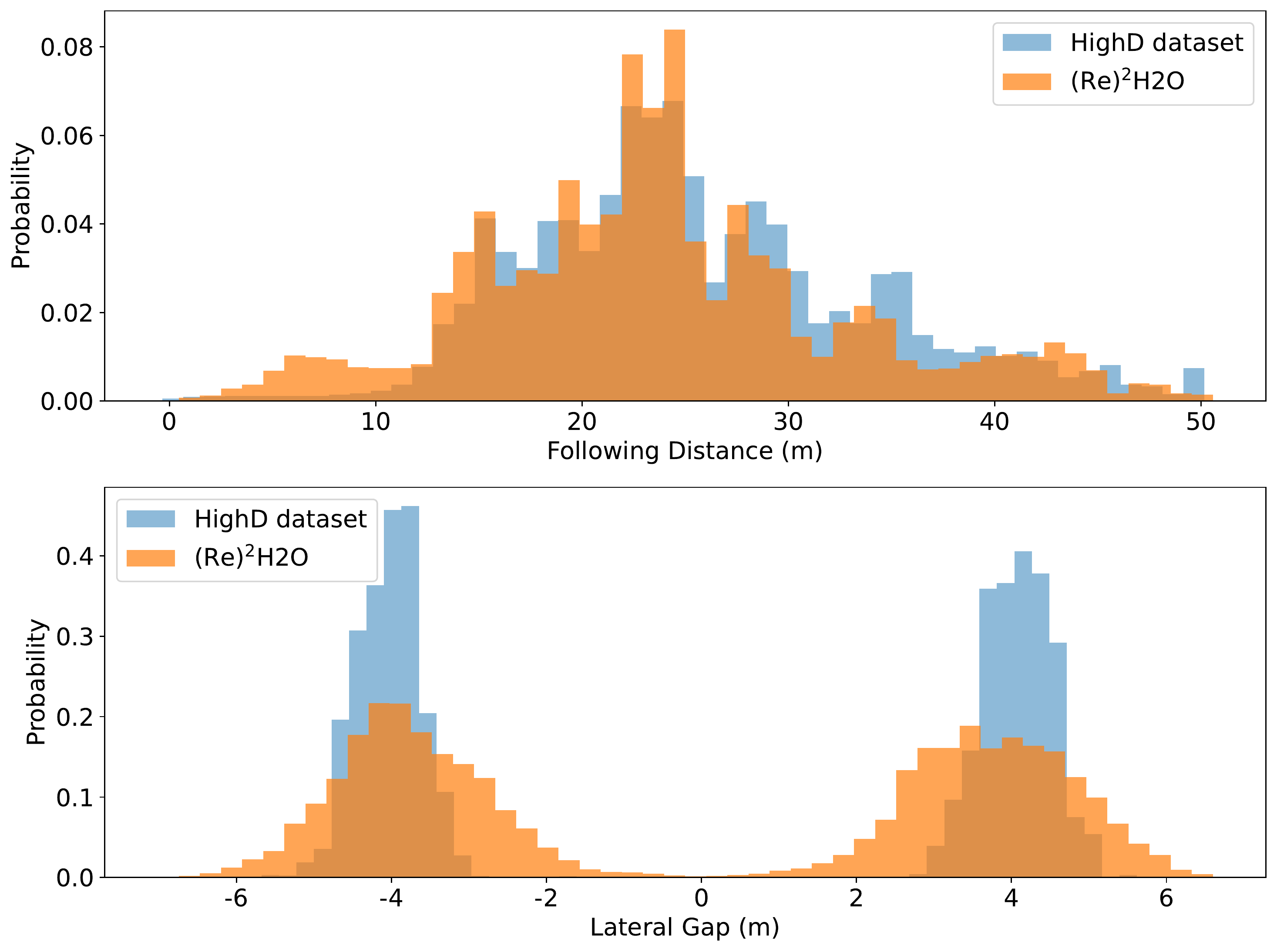}
    }
    \caption{Comparison between NDD and scenarios generated by (Re)$^2$H2O}
    \label{compare_to_HighD}
    \vspace{-5mm}
\end{figure}

\subsection{Experimental Results}\label{4-2}
\subsubsection{Scenario Analysis}\label{NDD}
In this paragraph, we analyze the scenario characteristics that differ from HighD and verify that our method is capable of generating more risky scenarios out of NDD. We initiate the state information of four BVs and one AV from processed five-vehicle scenarios from HighD, assign AV with ``fvdm" model and train BVs with (Re)$^2$H2O.


In Figure \ref{compare_to_HighD}, the distribution of following distance with front vehicle and lateral gap with parallel vehicle are visualized to compare the difference between NDD and generated scenarios.
Notably, (Re)$^2$H2O can generate scenarios with smaller following distance ($0\sim10$m), as well as more narrow lateral gap ($-3\sim3$m), which suggests that it produces scenarios with higher probability for accidents caused by both lane changing and vehicle following.

\begin{figure}
    \adjustbox{center}{
        \centering
        \begin{subfigure}{0.45\textwidth}
            \includegraphics[width=\textwidth]{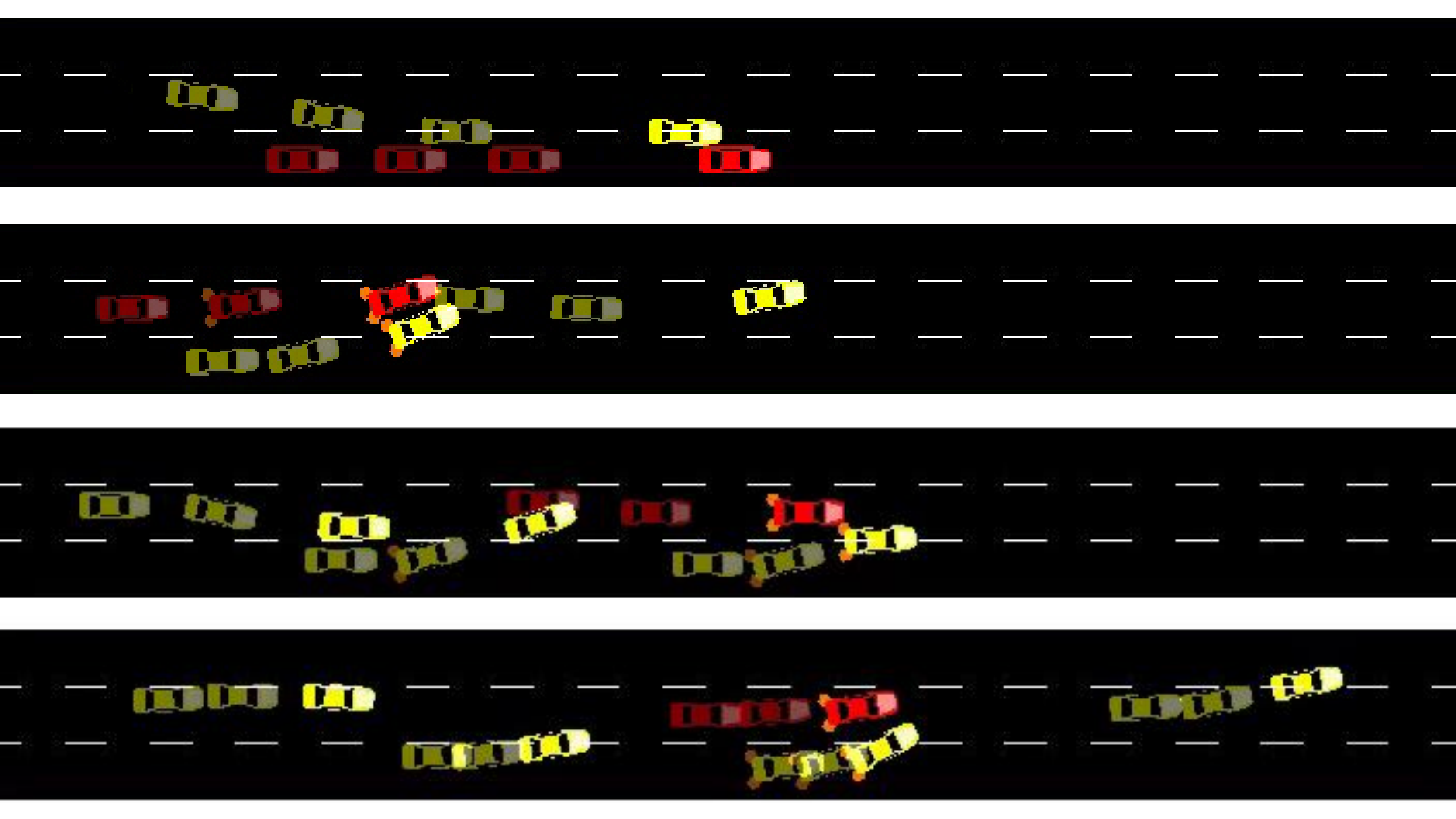}
        \end{subfigure}
    }
    \caption{Scenario generated by (Re)$^2$H2O with ``fvdm" AV model. AV is in red and BVs are in yellow. It should be noted that fvdm model is less sensitive to vehicles from the rear, so BVs find their best solution to follow AV  and perform rear collisions.}
    \label{sumo_scene}
    \vspace{-2mm}
\end{figure}

\begin{figure}
    \adjustbox{center}{
        \centering
            \includegraphics[width=0.5\textwidth]{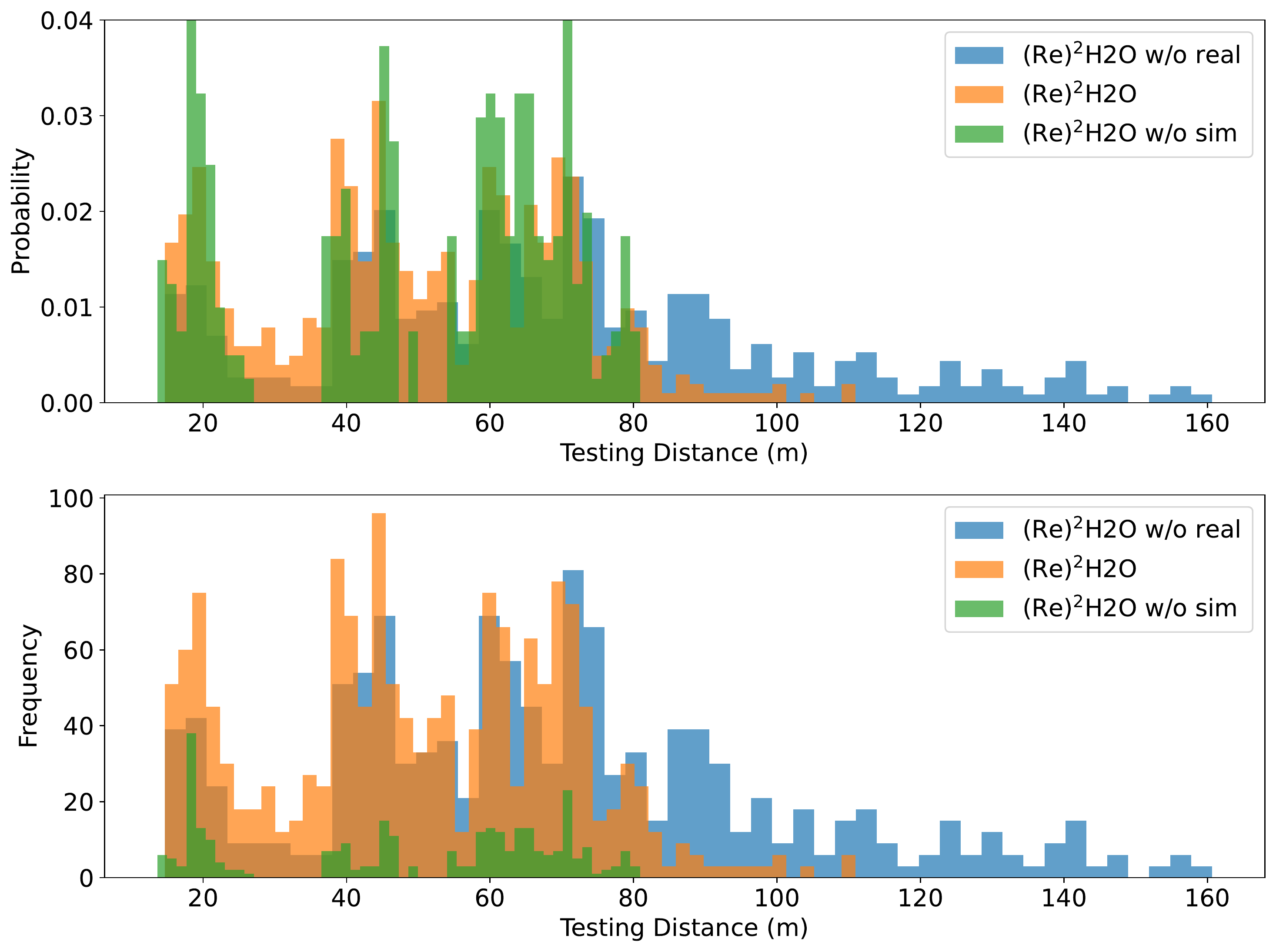}
    }
    \caption{Distance distribution of AV-BV collision scenarios. 
    High frequency of collisions on short distances may indicate efficient generation of risky scenarios, while high probability of collisions on short distances cannot support the conclusion, as the total number of collisions is unknown.
    }
    \label{baselines comparison}
    \vspace{-5mm}
\end{figure}


            
            
            

\begingroup
  \begin{table*}[t]
  \setlength{\tabcolsep}{4pt}
  \renewcommand{\arraystretch}{1.5}
  \tabcolsep=2pt
  \centering
  \caption{Scenario assessment against other generation baselines. Results are averaged over 3 random seeds.}
  \begin{threeparttable}
  \small
  \adjustbox{center}{
      \begin{tabular}{cccccccccccccccccccccc}
      \toprule
      \multirow{2}{*}{AV}   &  \multirow{2}{*}{BV Policy}             & \multicolumn{5}{c}{1 BV} & \multicolumn{5}{c}{2 BVs} & \multicolumn{5}{c}{3 BVs} & \multicolumn{5}{c}{4 BVs} \\ \cmidrule(l){3-7} \cmidrule(l){8-12} \cmidrule(l){13-17} \cmidrule(l){18-22}
         & & CR & ACT & ACD & CPS & CPM & CR & ACT & ACD & CPS & CPM  & CR &  ACT & ACD & CPS & CPM  & CR &  ACT & ACD & CPS & CPM   \\ \hline
      \multirow{4}{*}{\rotatebox{90}{uniform}}
            & \textbf{(Re)$^2$H2O}
                & 73.03  & 0.72  & 27.50  & \textbf{1.03} & \textbf{3.68} & 46.80  & 1.04  & 43.30  & \textbf{0.63} & \textbf{2.45} & 44.78  & 0.77  & 40.62  & \textbf{0.66} & \textbf{2.61} & 28.52  & 0.68  & 51.99  & \textbf{0.48} & \textbf{1.98} \\
            & \textit{(w/o real)}
                & \textbf{93.72} & 1.06  & 36.94  & 0.76  & 2.75  & \textbf{74.92} & 1.50  & 58.23  & 0.39  & 1.57  & 43.37  & 1.48  & 61.44  & 0.21  & 0.92  & 21.39  & 1.35  & 67.97  & 0.12  & 0.54  \\
            & \textit{(w/o sim)}
                & 24.11  & \textbf{0.65} & \textbf{18.25} & 0.25  & 0.87  & 4.86  & \textbf{0.57} & \textbf{28.38} & 0.09  & 0.35  & 8.29  & \textbf{0.47} & \textbf{26.56} & 0.16  & 0.63  & 7.00  & \textbf{0.45} & \textbf{37.76} & 0.15  & 0.60  \\\cline{2-22}
            & DR
                & 49.50  & 1.20  & 29.32  & 0.00  & 0.00  & 73.60  & 1.15  & 47.09  & 0.00  & 0.00  & \textbf{74.00} & 1.11  & 51.89  & 0.00  & 0.00  & \textbf{62.90} & 1.06  & 53.05  & 0.00  & 0.00  \\\hline
      \multirow{4}{*}{\rotatebox{90}{sumo}}
          & \textbf{(Re)$^2$H2O}
            & 68.95  & 0.64  & 25.02  & \textbf{1.06} & \textbf{3.67} & 36.40  & 1.05  & 42.65  & \textbf{0.53} & \textbf{1.99} & \textbf{38.37} & 0.70  & 37.98  & \textbf{0.58} & \textbf{2.24} & \textbf{24.80} & 0.59  & 48.84  & \textbf{0.43} & \textbf{1.72} \\
          & \textit{(w/o real)}
            & \textbf{79.25} & 0.75  & 27.97  & 0.58  & 1.75  & \textbf{53.14} & 1.37  & 51.05  & 0.23  & 0.82  & 30.84  & 1.31  & 54.96  & 0.14  & 0.53  & 20.44  & 1.22  & 67.65  & 0.12  & 0.47  \\
          & \textit{(w/o sim)}
            & 16.01  & \textbf{0.57} & \textbf{17.28} & 0.16  & 0.63  & 4.82  & \textbf{0.46} & \textbf{22.81} & 0.13  & 1.13  & 8.08  & \textbf{0.45} & \textbf{25.97} & 0.16  & 0.61  & 6.02  & \textbf{0.43} & \textbf{37.70} & 0.13  & 0.50  \\\cline{2-22}
          & DR
            &     0.00  & -     & -     & 0.00  & 0.00  & 0.00  & -     & -     & 0.00  & 0.00  & 0.00  & -     & -     & 0.00  & 0.00  & 0.00  & -     & -     & 0.00  & 0.00  \\\hline
      \multirow{4}{*}{\rotatebox{90}{fvdm}}
          & \textbf{(Re)$^2$H2O}
            & 75.48  & 0.66  & 26.02  & \textbf{1.09} & \textbf{3.63} & 41.97  & 1.05  & 43.60  & \textbf{0.51} & \textbf{1.81} & \textbf{38.46} & 0.62  & 37.12  & \textbf{0.65} & \textbf{2.41} & \textbf{25.13} & 0.62  & 50.11  & \textbf{0.49} & \textbf{1.90} \\
          & \textit{(w/o real)}
            & \textbf{79.06} & 0.67  & 26.24  & 0.60  & 1.77  & \textbf{57.22} & 1.38  & 53.03  & 0.26  & 0.86  & 35.46  & 1.29  & 55.82  & 0.17  & 0.61  & 21.27  & 1.18  & 68.38  & 0.11  & 0.39  \\
          & \textit{(w/o sim)}
            & 27.56  & \textbf{0.46} & \textbf{15.67} & 0.39  & 1.35  & 4.55  & \textbf{0.47} & \textbf{22.25} & 0.09  & 0.30  & 7.95  & \textbf{0.48} & \textbf{26.71} & 0.15  & 0.57  & 4.94  & \textbf{0.45} & \textbf{37.60} & 0.11  & 0.42  \\\cline{2-22}
          & DR
            & 17.10  & 2.58  & 41.66  & 0.00  & 0.00  & 27.90  & 2.20  & 60.44  & 0.00  & 0.00  & 24.40  & 1.94  & 61.67  & 0.00  & 0.00  & 1.00  & 2.37  & 66.86  & 0.00  & 0.00  \\   \bottomrule
      \end{tabular}\
   }
  \begin{tablenotes}
    \item[1] CR, ACT, ACD, CPS, and CPM are measured in units of percent, seconds, meters, times per second and times per 100 meter, respectively.
    \item[2] We expect higher values for CR, CPS and CPM, and lower values for ACT and ACD.
  \end{tablenotes}
  \end{threeparttable}
  \label{tab:main}
  \end{table*}
\endgroup

\subsubsection{Comparison Experiments}\label{compare}
In Table \ref{tab:main}, we evaluate (Re)$^2$H2O and multiple baselines with different AV Testing policies, varying the number of BVs. The collision scenes generated by (Re)$^2$H2O is snapshot in Figure \ref{sumo_scene}.
All the results are derived via exponential smoothing with a coefficient of 0.99 until the algorithmic convergence and averaged over 3 random seeds.

\begin{itemize}[leftmargin=*,topsep=0pt]
    \item \textbf{Comparisons on CR, ACT and ACD}: 
    Only in complex scenarios involving numerous BVs and more intelligent AV testing models does (Re)$^2$H2O achieve higher CR than baselines; in contrast, (Re)$^2$H2O \textit{w/o real} performs better in successfully invoking collisions in simpler situations. 
    Nonetheless, (Re)$^2$H2O \textit{w/o real} takes longer time duration and AV testing distance in each collision scenario.
    Meanwhile, (Re)$^2$H2O \textit{w/o sim} harvests all the top small value for ACT and ACD but obtains low CR, implicating that BV behaviors are conceivably equivalent to random actions.
    DR happens to be expert in accident generation with more BVs and na\"ive AV testing model, which unleashes the potential of randomization techniques to create chaos.
    However, it totally fails to generalize to risk ``sumo" AV driving model.
    To summarize, none of CR, ACT and ACD can provide clear evidence that one of the approached can cut down the time consumption and traveling distance for the AV testing pipeline.
    
    
    \item \textbf{Comparisons on CPS and CPM}: 
    To clearly demonstrate whether (Re)$^2$H2O generates risky scenarios and accidents with higher temporal and spatial efficiency, we compare (Re)$^2$H2O and baselines on CPS and CPM.
    (Re)$^2$H2O achieves the highest CPS and CPM in all the tasks and even wins by a twice to fivefold margin.
    It suggests that (Re)$^2$H2O is able to generate more collision cases in a certain testing time and AV distance and surely facilitate the testing procedure.
    We depict the probability and frequency distribution over testing distance of AV-BV collision cases in Figure \ref{baselines comparison}. 
    (Re)$^2$H2O obtains both high probability and frequency over small testing distances, while other baselines can only do well in one of either, e.g. (Re)$^2$H2O \textit{w/o sim} performs higher probability yet lower frequency at short AV-BV collision distances after initialization, which echoes the findings in Table~\ref{tab:main} that smaller ACD yet lower CR might also lead to extremely small CPM. 
    The employment of CPS and CPM considerately suit the widely-adopted augmented reality AV testing pipeline~\cite{liu2018real} that works as challenging the on-road AV model with simulated BVs controlled by generation algorithms.
    



\end{itemize}


\begin{figure}
    \adjustbox{center}{
        \centering
        \begin{subfigure}{0.195\textwidth}
            \caption{Average reward for AV}
            \includegraphics[width=\textwidth]{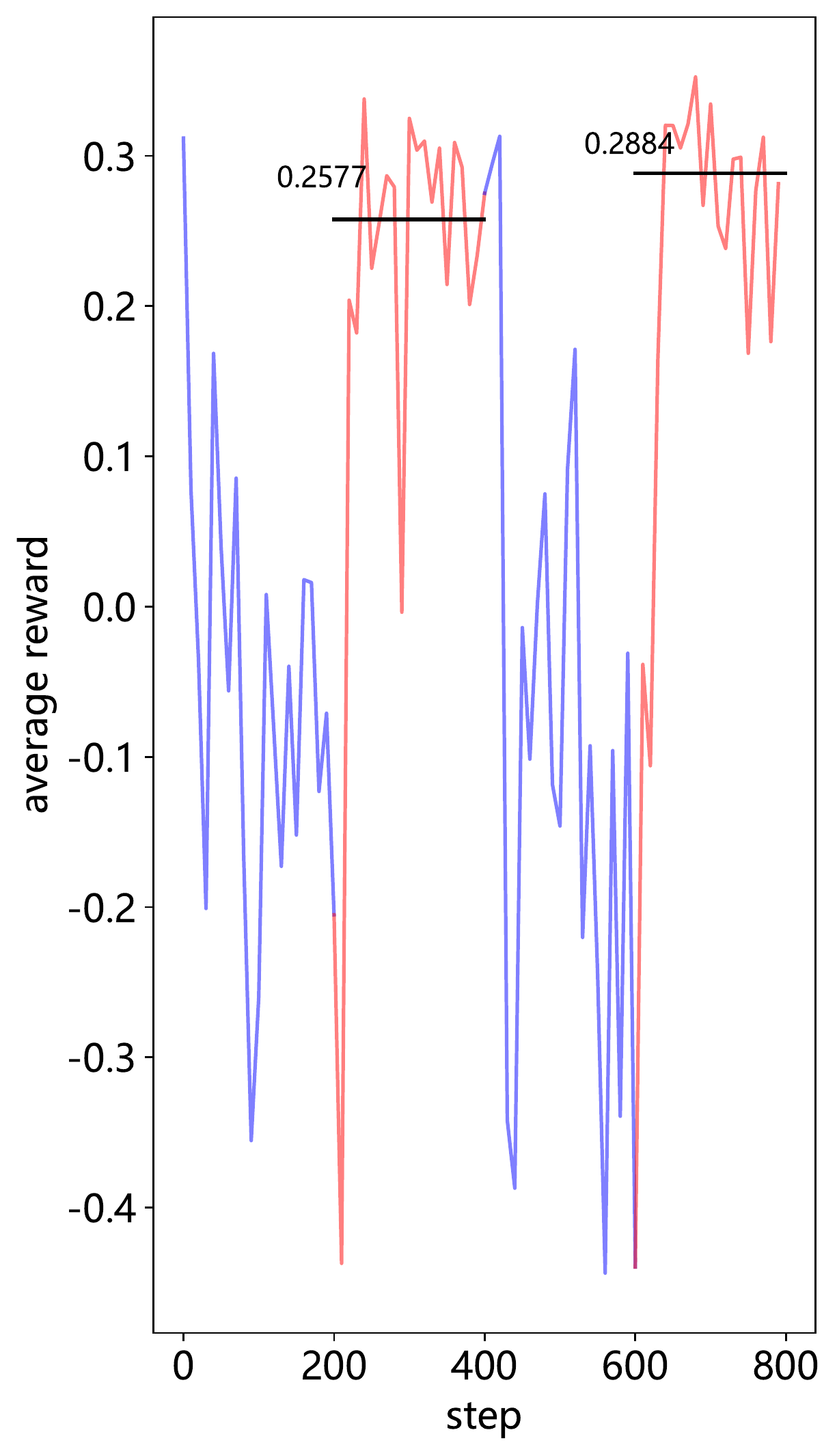} \label{AV reward}
        \end{subfigure}
        \begin{subfigure}{0.19\textwidth}
            \caption{Average speed for AV}
            \includegraphics[width=\textwidth]{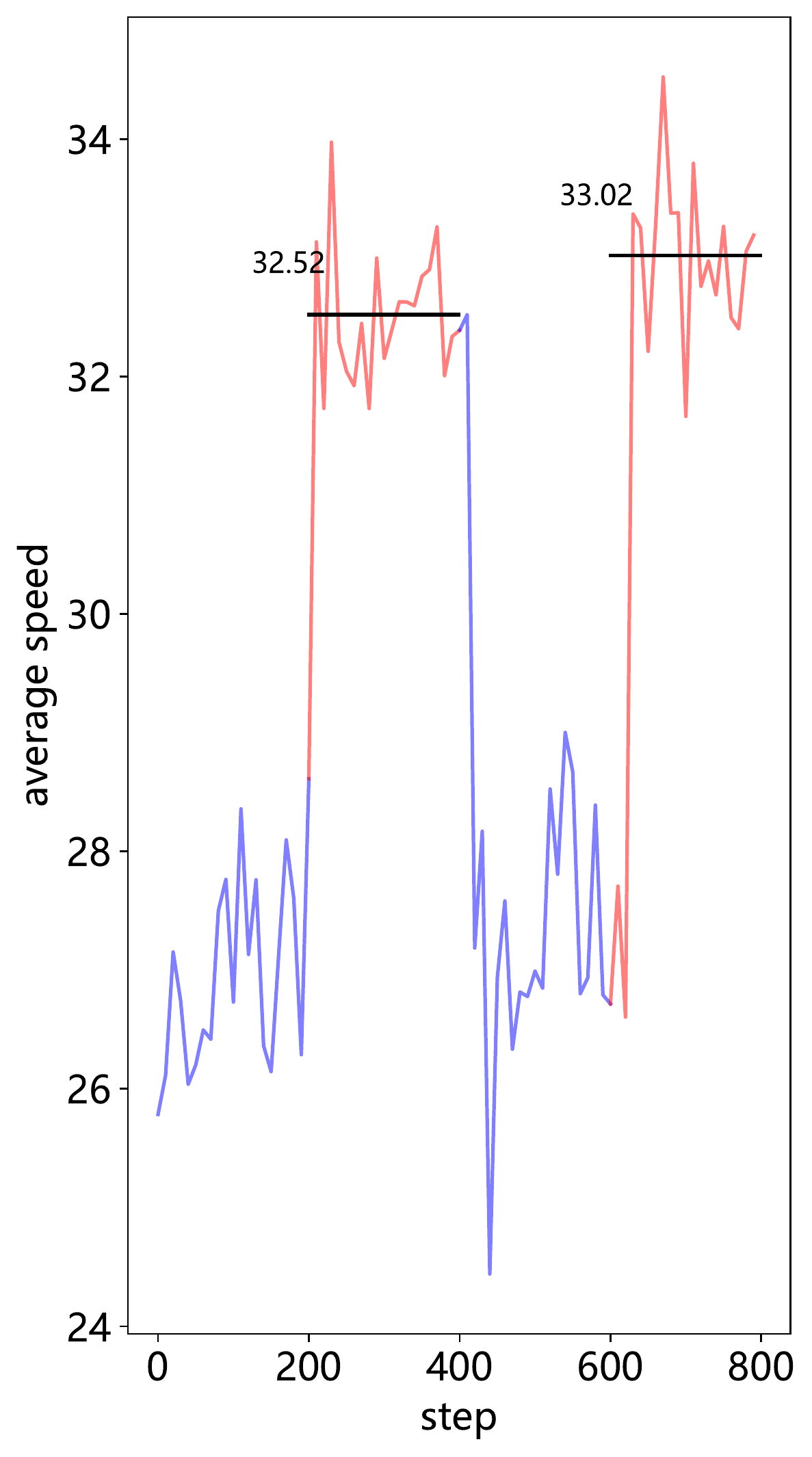} \label{AV speed}
        \end{subfigure}
    }
    \vspace{-5mm}
    \caption{Average rewards and speed for AV. Blue line represents BV training phase. Red line represents AV training phase. It is sufficient for AV to converge in each phase.}
    \label{fig:finetune}
    \vspace{-5mm}
\end{figure}

\subsubsection{Fine-tuning Experiment}\label{fine}
As mentioned above, (Re)$^2$H2O proves to generate risky scenarios more efficiently and accelerate AV testing and evaluation, while whether it also helps with AV performance improvement remains to be confirmed, which highly relates to the working efficacy of closed-loop AV industrial development.
Here we therefore evaluate whether an optimizable AV model can perform better by fine-tuning generated scenarios.
In the first phase, AV agent model is trained via Soft Actor Critic (SAC)\cite{haarnoja18b} algorithm in scenarios with 4 BVs equipped with ``fvdm" model.
The reward function for AV is designed as Eq.~\ref{reward function for AV}, encouraging higher velocity with a safety guarantee. 
In the second phase, BV policy is optimized with (Re)$^2$H2O and we fix the AV policy. At the same time, we evaluate AV model with scenarios generated by BVs.
Then in the next phase, we fine-tune AV policy on training scenarios augmented by (Re)$^2$H2O-based BV policy. Meanwhile, we always examine AV model with ``fvdm" BV policy.
With every phase lasting 200 training steps, we iterate AV and BV optimization to adversarially behave to each other and record the evaluated average reward and average speed of AV at each training step in Figure \ref{fig:finetune}.
\begin{equation}
r_{t,ego}=r_{t,ego}^{v}+r_{t,ego}^{col}
\label{reward function for AV}
\end{equation} 
\begin{equation}
r_{t,ego}^{v}=\frac{v-v_{max}/2}{v_{max}}
\end{equation}
\begin{equation}
r_{t,ego}^{col}=\left\{\begin{aligned}
-R_b,& \text{if AV collided with BV}
\\0,& \text{else}
\end{aligned}\right.
\end{equation}
In BV training phase, average reward and speed are pushed down since BVs keep learning more adversarial behaviors. 
In AV training phase, those values exceed themselves in previous AV training phase and converge in the end.
After fine-tuning AV policy for one turn, the average reward rises from 0.2577 to 0.2844, and the average speed rises from 32.5195 m/s to 33.0167 m/s, increasing by 1.53\%. 

We also carry out a similar experiment with (Re)$^2$H2O \textit{w/o real} for BV training, since it ranks only second to (Re)$^2$H2O on CPS and CPM. 
Compared with (Re)$^2$H2O, this approach gets higher average AV rewards in BV training phase, yet lower ones in AV training phase. 
More concretely, average AV speed is fine-tuned from 32.73 to 32.70 with no growth and collision rate rises from 6.23\% to 9\% 
This indicates that AV fine-tuned on scenarios generated by our methods can drive safer and more efficient, and not all scenario generation methods is capable of helping enhance AV performance. 
Furthermore, combining with the conclusions in Sec.~\ref{compare}, we validate that (Re)$^2$H2O can generalize to risk various rule-based AV driving models as well as learning-based ones.

\begin{figure*}[htpb]
    \centering
    \includegraphics[width=\textwidth]{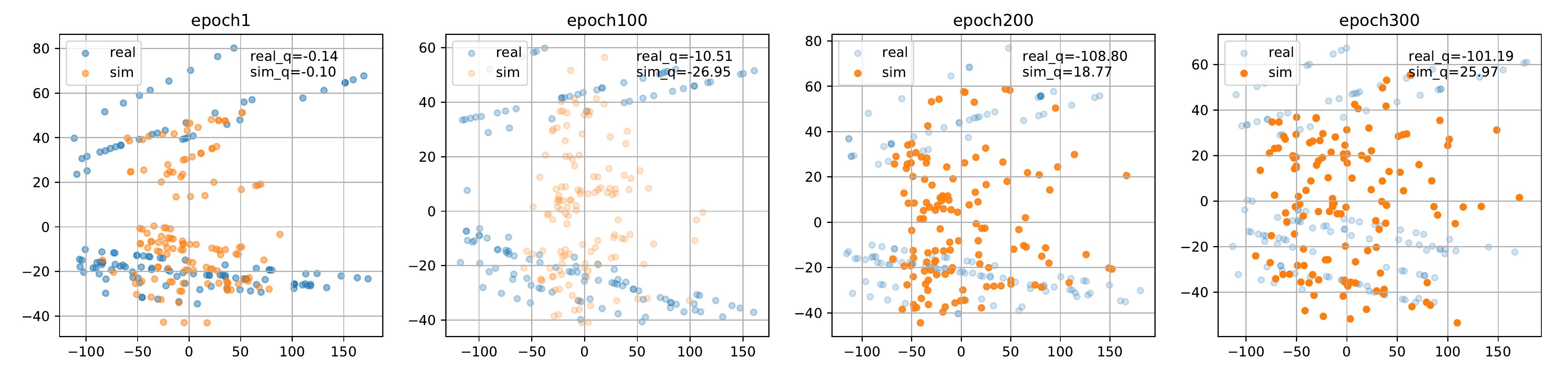} 
    \caption{An illustrative example explaining how (Re)$^2$H2O gradually learns to generate risky scenarios.}
    \label{pca}
    \vspace{-3mm}
\end{figure*}

\subsection{Illustrative Example}\label{didactic}
In order to demonstrate how (Re)$^2$H2O works to regularize Q-values on offline real data and online simulation data, we collect transitions in real and simulation batches at training epoch 1, 100, 200, and 300 until convergence.
In Figure \ref{pca}, we map the high-dimensional state-action pairs in samples into two dimensions with Principal Component Analysis (PCA).
We label the corresponding Q-values of samples with transparency. 
It is notable that NDD distribution is relatively narrow since naturalistic behaviors are overly safe thereby lacking in diversity, and the number of NDD data points used for training is limited to about 2000. 
On the other hand, with a replay buffer that contains $10^6$ data points, simulation can explore into a wider distribution of scenarios.
At the first beginning (epoch 1-100), Q-values on NDD are pushed down so that online simulation quickly explores wide distribution of driving scenarios, seeking more rewarding state-action combination.
With the value regularization mechanism illustrated in Eq.~\ref{new_pe}, (Re)$^2$H2O steadily pushes down Q-values on NDD and pulls up Q-values on simulation samples during epoch 100-300, resulting in highly-rewarded BV policy and thus generating more risky scenarios than NDD.


\section{Conclusion and Future Work}\label{5}
In this paper, we propose a novel recipe for autonomous driving scenario generation that learns adversarial BV policies from both offline NDD and online simulation samples, dubbing it (Re)$^2$H2O. 
Through hybrid offline-and-online RL with a reverse value regularization scheme, Q-values on NDD samples are penalized while those on simulation rollouts are boosted, encouraging BV to explore varied and aggressive driving behaviors and also cutting down the search space with NDD as implicit constraints.
In contrast to other competing baselines, (Re)$^2$H2O can produce risky and diverse scenarios from naturalistic driving datasets with overwhelmingly higher temporal and spatial efficiency, which would radically accelerate AV testing procedures.
In addition, (Re)$^2$H2O can generalize to challenge various autonomous driving models and these generated scenarios are also corroborated to be capable of fine-tuning optimizable AV policies.
We highlight the significance and superiority of (Re)$^2$H2O that potentially sheds light on the development of testing case generation methods for all the safety-critical intelligent decision-making systems.
In the future, we are interested in experimenting with more intricate road topologies and conditions as well as a broader choice of background traffic participants.



\section*{Acknowledgement}
This work is supported by National Natural Science Foundation of China under Grant No.~62133002. Haoyi Niu is also funded by Tsinghua Undergraduate ``Future Scholar'' Scientific Research Grant, i.e. Tsinghua University Initiative Scientific Research Program (20217020007). 


\bibliographystyle{IEEEtran}
\bibliography{mylib}

\begin{thebibliography}{10}
\providecommand{\url}[1]{#1}
\csname url@rmstyle\endcsname
\providecommand{\newblock}{\relax}
\providecommand{\bibinfo}[2]{#2}
\providecommand\BIBentrySTDinterwordspacing{\spaceskip=0pt\relax}
\providecommand\BIBentryALTinterwordstretchfactor{4}
\providecommand\BIBentryALTinterwordspacing{\spaceskip=\fontdimen2\font plus
\BIBentryALTinterwordstretchfactor\fontdimen3\font minus
  \fontdimen4\font\relax}
\providecommand\BIBforeignlanguage[2]{{%
\expandafter\ifx\csname l@#1\endcsname\relax
\typeout{** WARNING: IEEEtran.bst: No hyphenation pattern has been}%
\typeout{** loaded for the language `#1'. Using the pattern for}%
\typeout{** the default language instead.}%
\else
\language=\csname l@#1\endcsname
\fi
#2}}

\bibitem{filos2020can}
A.~Filos, P.~Tigkas, R.~McAllister, N.~Rhinehart, S.~Levine, and Y.~Gal, ``Can
  autonomous vehicles identify, recover from, and adapt to distribution
  shifts?'' in \emph{International Conference on Machine Learning}.\hskip 1em
  plus 0.5em minus 0.4em\relax PMLR, 2020, pp. 3145--3153.

\bibitem{kalra2016driving}
N.~Kalra and S.~M. Paddock, ``Driving to safety: How many miles of driving
  would it take to demonstrate autonomous vehicle reliability?''
  \emph{Transportation Research Part A: Policy and Practice}, vol.~94, pp.
  182--193, 2016.

\bibitem{li2016intelligence}
L.~Li, W.-L. Huang, Y.~Liu, N.-N. Zheng, and F.-Y. Wang, ``Intelligence testing
  for autonomous vehicles: A new approach,'' \emph{IEEE Transactions on
  Intelligent Vehicles}, vol.~1, no.~2, pp. 158--166, 2016.

\bibitem{li2018artificial}
L.~Li, Y.-L. Lin, N.-N. Zheng, F.-Y. Wang, Y.~Liu, D.~Cao, K.~Wang, and W.-L.
  Huang, ``Artificial intelligence test: A case study of intelligent
  vehicles,'' \emph{Artificial Intelligence Review}, vol.~50, no.~3, pp.
  441--465, 2018.

\bibitem{li2019parallel}
L.~Li, X.~Wang, K.~Wang, Y.~Lin, J.~Xin, L.~Chen, L.~Xu, B.~Tian, Y.~Ai,
  J.~Wang, \emph{et~al.}, ``Parallel testing of vehicle intelligence via
  virtual-real interaction,'' \emph{Science robotics}, vol.~4, no.~28, p.
  eaaw4106, 2019.

\bibitem{li2020theoretical}
L.~Li, N.~Zheng, and F.-Y. Wang, ``A theoretical foundation of intelligence
  testing and its application for intelligent vehicles,'' \emph{IEEE
  Transactions on Intelligent Transportation Systems}, vol.~22, no.~10, pp.
  6297--6306, 2020.

\bibitem{feng2021intelligent}
S.~Feng, X.~Yan, H.~Sun, Y.~Feng, and H.~X. Liu, ``Intelligent driving
  intelligence test for autonomous vehicles with naturalistic and adversarial
  environment,'' \emph{Nature communications}, vol.~12, no.~1, pp. 1--14, 2021.

\bibitem{zhao2016accelerated}
D.~Zhao, ``Accelerated evaluation of automated vehicles.'' Ph.D. dissertation,
  2016.

\bibitem{o2018scalable}
M.~O'Kelly, A.~Sinha, H.~Namkoong, R.~Tedrake, and J.~C. Duchi, ``Scalable
  end-to-end autonomous vehicle testing via rare-event simulation,''
  \emph{Advances in neural information processing systems}, vol.~31, 2018.

\bibitem{xusafebench}
C.~Xu, W.~Ding, W.~Lyu, Z.~Liu, S.~Wang, Y.~He, H.~Hu, D.~Zhao, and B.~Li,
  ``Safebench: A benchmarking platform for safety evaluation of autonomous
  vehicles,'' in \emph{Thirty-sixth Conference on Neural Information Processing
  Systems Datasets and Benchmarks Track}.

\bibitem{ding2022survey}
W.~Ding, C.~Xu, H.~Lin, B.~Li, and D.~Zhao, ``A survey on safety-critical
  scenario generation from methodological perspective,'' \emph{arXiv e-prints},
  pp. arXiv--2202, 2022.

\bibitem{kruber2018unsupervised}
F.~Kruber, J.~Wurst, and M.~Botsch, ``An unsupervised random forest clustering
  technique for automatic traffic scenario categorization,'' in \emph{2018 21st
  International conference on intelligent transportation systems (ITSC)}.\hskip
  1em plus 0.5em minus 0.4em\relax IEEE, 2018, pp. 2811--2818.

\bibitem{wang2018extracting}
W.~Wang and D.~Zhao, ``Extracting traffic primitives directly from
  naturalistically logged data for self-driving applications,'' \emph{IEEE
  Robotics and Automation Letters}, vol.~3, no.~2, pp. 1223--1229, 2018.

\bibitem{scanlon2021waymo}
J.~M. Scanlon, K.~D. Kusano, T.~Daniel, C.~Alderson, A.~Ogle, and T.~Victor,
  ``Waymo simulated driving behavior in reconstructed fatal crashes within an
  autonomous vehicle operating domain,'' \emph{Accident Analysis \&
  Prevention}, vol. 163, p. 106454, 2021.

\bibitem{wheeler2015initial}
T.~A. Wheeler, M.~J. Kochenderfer, and P.~Robbel, ``Initial scene
  configurations for highway traffic propagation,'' in \emph{2015 IEEE 18th
  International Conference on Intelligent Transportation Systems}.\hskip 1em
  plus 0.5em minus 0.4em\relax IEEE, 2015, pp. 279--284.

\bibitem{huang2018synthesis}
Z.~Huang, M.~Arief, H.~Lam, and D.~Zhao, ``Synthesis of different autonomous
  vehicles test approaches,'' in \emph{2018 21st International Conference on
  Intelligent Transportation Systems (ITSC)}.\hskip 1em plus 0.5em minus
  0.4em\relax IEEE, 2018, pp. 2000--2005.

\bibitem{sun2021corner}
H.~Sun, S.~Feng, X.~Yan, and H.~X. Liu, ``Corner case generation and analysis
  for safety assessment of autonomous vehicles,'' \emph{Transportation research
  record}, vol. 2675, no.~11, pp. 587--600, 2021.

\bibitem{akagi2019risk}
Y.~Akagi, R.~Kato, S.~Kitajima, J.~Antona-Makoshi, and N.~Uchida, ``A
  risk-index based sampling method to generate scenarios for the evaluation of
  automated driving vehicle safety,'' in \emph{2019 IEEE Intelligent
  Transportation Systems Conference (ITSC)}.\hskip 1em plus 0.5em minus
  0.4em\relax IEEE, 2019, pp. 667--672.

\bibitem{koren2018adaptive}
M.~Koren, S.~Alsaif, R.~Lee, and M.~J. Kochenderfer, ``Adaptive stress testing
  for autonomous vehicles,'' in \emph{2018 IEEE Intelligent Vehicles Symposium
  (IV)}.\hskip 1em plus 0.5em minus 0.4em\relax IEEE, 2018, pp. 1--7.

\bibitem{koren2019efficient}
M.~Koren and M.~J. Kochenderfer, ``Efficient autonomy validation in simulation
  with adaptive stress testing,'' in \emph{2019 IEEE Intelligent Transportation
  Systems Conference (ITSC)}.\hskip 1em plus 0.5em minus 0.4em\relax IEEE,
  2019, pp. 4178--4183.

\bibitem{lee2020adaptive}
R.~Lee, O.~J. Mengshoel, A.~Saksena, R.~W. Gardner, D.~Genin, J.~Silbermann,
  M.~Owen, and M.~J. Kochenderfer, ``Adaptive stress testing: Finding likely
  failure events with reinforcement learning,'' \emph{Journal of Artificial
  Intelligence Research}, vol.~69, pp. 1165--1201, 2020.

\bibitem{corso2019adaptive}
A.~Corso, P.~Du, K.~Driggs-Campbell, and M.~J. Kochenderfer, ``Adaptive stress
  testing with reward augmentation for autonomous vehicle validatio,'' in
  \emph{2019 IEEE Intelligent Transportation Systems Conference (ITSC)}.\hskip
  1em plus 0.5em minus 0.4em\relax IEEE, 2019, pp. 163--168.

\bibitem{songhybrid}
Y.~Song, Y.~Zhou, A.~Sekhari, D.~Bagnell, A.~Krishnamurthy, and W.~Sun,
  ``Hybrid rl: Using both offline and online data can make rl efficient,'' in
  \emph{3rd Offline RL Workshop: Offline RL as a''Launchpad''}.

\bibitem{wagenmaker2022leveraging}
A.~Wagenmaker and A.~Pacchiano, ``Leveraging offline data in online
  reinforcement learning,'' \emph{arXiv preprint arXiv:2211.04974}, 2022.

\bibitem{niu2022when}
\BIBentryALTinterwordspacing
H.~Niu, S.~Sharma, Y.~Qiu, M.~Li, G.~Zhou, J.~HU, and X.~Zhan, ``When to trust
  your simulator: Dynamics-aware hybrid offline-and-online reinforcement
  learning,'' in \emph{Advances in Neural Information Processing Systems},
  2022. [Online]. Available: \url{https://openreview.net/forum?id=zXE8iFOZKw}
\BIBentrySTDinterwordspacing

\bibitem{krajewski2018highd}
R.~Krajewski, J.~Bock, L.~Kloeker, and L.~Eckstein, ``The highd dataset: A
  drone dataset of naturalistic vehicle trajectories on german highways for
  validation of highly automated driving systems,'' in \emph{2018 21st
  International Conference on Intelligent Transportation Systems (ITSC)}.\hskip
  1em plus 0.5em minus 0.4em\relax IEEE, 2018, pp. 2118--2125.

\bibitem{behrisch2011sumo}
M.~Behrisch, L.~Bieker, J.~Erdmann, and D.~Krajzewicz, ``Sumo--simulation of
  urban mobility: an overview,'' in \emph{Proceedings of SIMUL 2011, The Third
  International Conference on Advances in System Simulation}.\hskip 1em plus
  0.5em minus 0.4em\relax ThinkMind, 2011.

\bibitem{yang2020surfelgan}
Z.~Yang, Y.~Chai, D.~Anguelov, Y.~Zhou, P.~Sun, D.~Erhan, S.~Rafferty, and
  H.~Kretzschmar, ``Surfelgan: Synthesizing realistic sensor data for
  autonomous driving,'' in \emph{Proceedings of the IEEE/CVF Conference on
  Computer Vision and Pattern Recognition}, 2020, pp. 11\,118--11\,127.

\bibitem{chen2021geosim}
Y.~Chen, F.~Rong, S.~Duggal, S.~Wang, X.~Yan, S.~Manivasagam, S.~Xue, E.~Yumer,
  and R.~Urtasun, ``Geosim: Realistic video simulation via geometry-aware
  composition for self-driving,'' in \emph{Proceedings of the IEEE/CVF
  conference on computer vision and pattern recognition}, 2021, pp. 7230--7240.

\bibitem{ehrhardt2020relate}
S.~Ehrhardt, O.~Groth, A.~Monszpart, M.~Engelcke, I.~Posner, N.~Mitra, and
  A.~Vedaldi, ``Relate: Physically plausible multi-object scene synthesis using
  structured latent spaces,'' \emph{Advances in Neural Information Processing
  Systems}, vol.~33, pp. 11\,202--11\,213, 2020.

\bibitem{Mhakansson2021}
J.~W. M.Hakansson, ``Driving scenario generation using generative adversarial
  networks,'' Master's thesis, 2021.

\bibitem{chen2021adversarial}
B.~Chen, X.~Chen, Q.~Wu, and L.~Li, ``Adversarial evaluation of autonomous
  vehicles in lane-change scenarios,'' \emph{IEEE Transactions on Intelligent
  Transportation Systems}, vol.~23, no.~8, pp. 10\,333--10\,342, 2021.

\bibitem{ding2020learning}
W.~Ding, B.~Chen, M.~Xu, and D.~Zhao, ``Learning to collide: An adaptive
  safety-critical scenarios generating method,'' in \emph{2020 IEEE/RSJ
  International Conference on Intelligent Robots and Systems (IROS)}.\hskip 1em
  plus 0.5em minus 0.4em\relax IEEE, 2020, pp. 2243--2250.

\bibitem{wachi2019failure}
A.~Wachi, ``Failure-scenario maker for rule-based agent using multi-agent
  adversarial reinforcement learning and its application to autonomous
  driving,'' in \emph{IJCAI}, 2019.

\bibitem{abeysirigoonawardena2019generating}
Y.~Abeysirigoonawardena, F.~Shkurti, and G.~Dudek, ``Generating adversarial
  driving scenarios in high-fidelity simulators,'' in \emph{2019 International
  Conference on Robotics and Automation (ICRA)}.\hskip 1em plus 0.5em minus
  0.4em\relax IEEE, 2019, pp. 8271--8277.

\bibitem{klischat2019generating}
M.~Klischat and M.~Althoff, ``Generating critical test scenarios for automated
  vehicles with evolutionary algorithms,'' in \emph{2019 IEEE Intelligent
  Vehicles Symposium (IV)}.\hskip 1em plus 0.5em minus 0.4em\relax IEEE, 2019,
  pp. 2352--2358.

\bibitem{Ding2018ANM}
W.~Ding, W.~Wang, and D.~Zhao, ``A new multi-vehicle trajectory generator to
  simulate vehicle-to-vehicle encounters,'' \emph{arXiv: Computer Vision and
  Pattern Recognition}, 2018.

\bibitem{wang2021advsim}
J.~Wang, A.~Pun, J.~Tu, S.~Manivasagam, A.~Sadat, S.~Casas, M.~Ren, and
  R.~Urtasun, ``Advsim: Generating safety-critical scenarios for self-driving
  vehicles,'' in \emph{Proceedings of the IEEE/CVF Conference on Computer
  Vision and Pattern Recognition}, 2021, pp. 9909--9918.

\bibitem{hanselmann2022king}
N.~Hanselmann, K.~Renz, K.~Chitta, A.~Bhattacharyya, and A.~Geiger, ``King:
  Generating safety-critical driving scenarios for robust imitation via
  kinematics gradients,'' in \emph{Computer Vision--ECCV 2022: 17th European
  Conference, Tel Aviv, Israel, October 23--27, 2022, Proceedings, Part
  XXXVIII}.\hskip 1em plus 0.5em minus 0.4em\relax Springer, 2022, pp.
  335--352.

\bibitem{karunakaran2020efficient}
D.~Karunakaran, S.~Worrall, and E.~Nebot, ``Efficient statistical validation
  with edge cases to evaluate highly automated vehicles,'' in \emph{2020 IEEE
  23rd International Conference on Intelligent Transportation Systems
  (ITSC)}.\hskip 1em plus 0.5em minus 0.4em\relax IEEE, 2020, pp. 1--8.

\bibitem{rana2021building}
A.~Rana and A.~Malhi, ``Building safer autonomous agents by leveraging risky
  driving behavior knowledge,'' in \emph{2021 International Conference on
  Communications, Computing, Cybersecurity, and Informatics (CCCI)}.\hskip 1em
  plus 0.5em minus 0.4em\relax IEEE, 2021, pp. 1--6.

\bibitem{bagschik2018ontology}
G.~Bagschik, T.~Menzel, and M.~Maurer, ``Ontology based scene creation for the
  development of automated vehicles,'' in \emph{2018 IEEE Intelligent Vehicles
  Symposium (IV)}.\hskip 1em plus 0.5em minus 0.4em\relax IEEE, 2018, pp.
  1813--1820.

\bibitem{menzel2018scenarios}
T.~Menzel, G.~Bagschik, and M.~Maurer, ``Scenarios for development, test and
  validation of automated vehicles,'' in \emph{2018 IEEE Intelligent Vehicles
  Symposium (IV)}.\hskip 1em plus 0.5em minus 0.4em\relax IEEE, 2018, pp.
  1821--1827.

\bibitem{shiroshita2020behaviorally}
S.~Shiroshita, S.~Maruyama, D.~Nishiyama, M.~Y. Castro, K.~Hamzaoui, G.~Rosman,
  J.~DeCastro, K.-H. Lee, and A.~Gaidon, ``Behaviorally diverse traffic
  simulation via reinforcement learning,'' in \emph{2020 IEEE/RSJ International
  Conference on Intelligent Robots and Systems (IROS)}.\hskip 1em plus 0.5em
  minus 0.4em\relax IEEE, 2020, pp. 2103--2110.

\bibitem{althoff2018automatic}
M.~Althoff and S.~Lutz, ``Automatic generation of safety-critical test
  scenarios for collision avoidance of road vehicles,'' in \emph{2018 IEEE
  Intelligent Vehicles Symposium (IV)}.\hskip 1em plus 0.5em minus 0.4em\relax
  IEEE, 2018, pp. 1326--1333.

\bibitem{suo2021trafficsim}
S.~Suo, S.~Regalado, S.~Casas, and R.~Urtasun, ``Trafficsim: Learning to
  simulate realistic multi-agent behaviors,'' in \emph{Proceedings of the
  IEEE/CVF Conference on Computer Vision and Pattern Recognition}, 2021, pp.
  10\,400--10\,409.

\bibitem{ding2021multimodal}
W.~Ding, B.~Chen, B.~Li, K.~J. Eun, and D.~Zhao, ``Multimodal safety-critical
  scenarios generation for decision-making algorithms evaluation,'' \emph{IEEE
  Robotics and Automation Letters}, vol.~6, no.~2, pp. 1551--1558, 2021.

\bibitem{wheeler2016factor}
T.~A. Wheeler and M.~J. Kochenderfer, ``Factor graph scene distributions for
  automotive safety analysis,'' in \emph{2016 IEEE 19th International
  Conference on Intelligent Transportation Systems (ITSC)}.\hskip 1em plus
  0.5em minus 0.4em\relax IEEE, 2016, pp. 1035--1040.

\bibitem{knies2020data}
C.~Knies and F.~Diermeyer, ``Data-driven test scenario generation for
  cooperative maneuver planning on highways,'' \emph{Applied Sciences},
  vol.~10, no.~22, p. 8154, 2020.

\bibitem{ulbrich2015defining}
S.~Ulbrich, T.~Menzel, A.~Reschka, F.~Schuldt, and M.~Maurer, ``Defining and
  substantiating the terms scene, situation, and scenario for automated
  driving,'' in \emph{2015 IEEE 18th international conference on intelligent
  transportation systems}.\hskip 1em plus 0.5em minus 0.4em\relax IEEE, 2015,
  pp. 982--988.

\bibitem{feng2020testing1}
S.~Feng, Y.~Feng, C.~Yu, Y.~Zhang, and H.~X. Liu, ``Testing scenario library
  generation for connected and automated vehicles, part i: Methodology,''
  \emph{IEEE Transactions on Intelligent Transportation Systems}, vol.~22,
  no.~3, pp. 1573--1582, 2020.

\bibitem{sutton1998introduction}
R.~S. Sutton, A.~G. Barto, \emph{et~al.}, \emph{Introduction to reinforcement
  learning}.\hskip 1em plus 0.5em minus 0.4em\relax MIT press Cambridge, 1998,
  vol. 135.

\bibitem{niu2021dr2l}
H.~Niu, J.~Hu, Z.~Cui, and Y.~Zhang, ``Dr2l: Surfacing corner cases to
  robustify autonomous driving via domain randomization reinforcement
  learning,'' in \emph{Proceedings of the 5th International Conference on
  Computer Science and Application Engineering}, 2021, pp. 1--8.

\bibitem{song2014research}
J.~Song, Y.~Wu, Z.~Xu, and X.~Lin, ``Research on car-following model based on
  sumo,'' in \emph{The 7th IEEE/International Conference on Advanced Infocomm
  Technology}.\hskip 1em plus 0.5em minus 0.4em\relax IEEE, 2014, pp. 47--55.

\bibitem{erdmann2015sumo}
J.~Erdmann, ``Sumo’s lane-changing model,'' in \emph{Modeling Mobility with
  Open Data: 2nd SUMO Conference 2014 Berlin, Germany, May 15-16, 2014}.\hskip
  1em plus 0.5em minus 0.4em\relax Springer, 2015, pp. 105--123.

\bibitem{jiang2001full}
R.~Jiang, Q.~Wu, and Z.~Zhu, ``Full velocity difference model for a
  car-following theory,'' \emph{Physical Review E}, vol.~64, no.~1, p. 017101,
  2001.

\bibitem{liu2018real}
H.~Liu and Y.~Feng, ``Real world meets virtual world: Augmented reality makes
  driverless vehicle testing faster, safer, and cheaper,'' \emph{Mcity White
  Paper}, 2018.

\bibitem{haarnoja18b}
\BIBentryALTinterwordspacing
T.~Haarnoja, A.~Zhou, P.~Abbeel, and S.~Levine, ``Soft actor-critic: Off-policy
  maximum entropy deep reinforcement learning with a stochastic actor,'' in
  \emph{Proceedings of the 35th International Conference on Machine Learning},
  ser. Proceedings of Machine Learning Research, J.~Dy and A.~Krause, Eds.,
  vol.~80.\hskip 1em plus 0.5em minus 0.4em\relax PMLR, 10--15 Jul 2018, pp.
  1861--1870. [Online]. Available:
  \url{http://proceedings.mlr.press/v80/haarnoja18b.html}
\BIBentrySTDinterwordspacing

\end{thebibliography}

\end{document}